\documentclass[letterpaper, 10 pt, conference]{ieeeconf}  
\IEEEoverridecommandlockouts                              

\usepackage[utf8]{inputenc} 
\usepackage{graphics}   
\usepackage{amssymb, amsmath}   

\usepackage{amsthm}     
\usepackage{microtype}  
\usepackage{epstopdf}   
\usepackage{url}    
\usepackage[noadjust]{cite} 
\usepackage{overpic}    
\usepackage[linesnumbered,ruled,noend]{algorithm2e}    
\usepackage[dvipsnames]{xcolor} 
\usepackage{subfig}
\usepackage{wrapfig}
\usepackage[font=small,labelfont=bf]{caption}
\usepackage{multirow}   
\usepackage{titlesec}   
\usepackage{footnote}   
\usepackage{soul}   
\usepackage{tikz}   
\usetikzlibrary{tikzmark}
\usepackage{comment}
\usepackage[textsize=scriptsize]{todonotes}

\makeatletter
\let\NAT@parse\undefined
\makeatother
\usepackage{hyperref}   

\usepackage{etoolbox}

\makeatletter
\patchcmd{\@algocf@start}
  {-1.5em}
  {0pt}
  {}{}
\makeatother
 
\titlespacing\section{0pt}{6pt plus 2pt minus 2pt}{2pt plus 2pt minus 2pt}
\titlespacing\subsection{0pt}{4pt plus 2pt minus 2pt}{2pt plus 2pt minus 2pt}
\newtheorem{problem}{Problem}

\newtheorem{lemma}{Lemma}[section]

\newtheorem{theorem}{Theorem}[section]
\theoremstyle{definition}

\theoremstyle{remark}
\newtheorem{remark}{Remark}

\DeclareMathOperator*{\argmax}{arg\,max}

\newcommand{\suo}{\textsc{SUO}\xspace}
\newcommand{\suoi}{\textsc{SU-I}\xspace}

\newcommand{\spp}{\texttt{SPP}\xspace}
\newcommand{\mpp}{\texttt{MPP}\xspace}
\newcommand{\mppl}{\texttt{LMPP}\xspace}
\newcommand{\mapf}{\texttt{MAPF}\xspace}

\SetCommentSty{mycommfont}

\title{
Optimizing Space Utilization for More Effective Multi-Robot Path Planning
}
\author{Shuai D. Han$^{1}$ 
\quad Jingjin Yu$^{1}$
\thanks{
$^{1}$S. D. Han and J. Yu are with the Department of Computer Science, 
Rutgers University.
Emails: {\tt\small \{shuai.han, jingjin.yu\}@rutgers.edu}.
}%
}

\begin{document}

\maketitle
\thispagestyle{empty}
\pagestyle{empty}

\begin{abstract}
We perform a systematic exploration of the principle of Space Utilization 
Optimization (\suo) as a heuristic for planning better individual paths 
in a decoupled multi-robot path planner, with applications to both one-shot
and life-long multi-robot path planning problems. We show that the decentralized
heuristic set, \suoi, preserves single path optimality and significantly 
reduces congestion that naturally happens when many paths are planned without 
coordination. Integration of \suoi into complete planners brings 
dramatic reductions in computation time due to the significantly reduced number 
of conflicts and leads to sizable solution optimality gains in 
diverse evaluation scenarios with medium and large maps, for both one-shot 
and life-long problem settings. 
\end{abstract}

\section{Introduction}
Recent years have witnessed a dramatic acceleration in the deployment of 
multi-robot systems for general logistic tasks~\cite{dekhne2019automation}, 
especially in the domain of shipping and warehousing~\cite{WurDanMou08}. 
Fast-paced expansion is predicted across the board, with the warehouse 
domain alone expecting a $14\%$ year-over-year growth in the next five 
years~\cite{covid-auto}. 
This in turn demands the push for enhancing the scalability of multi-agent 
and multi-robot systems, which in the end boils down to achieving the 
maximum possible output attainable. Holding other variables constant, 
maximizing system throughput is most readily achieved by increasing robot 
density and plan optimality, which calls for faster and better computational 
methods for Multi-Robot Path Planning (\mpp) and Life-long Multi-Robot Path 
Planning (\mppl) problems.

Toward the development of more efficient and higher performance multi-robot systems
catering to the rapidly growing need of automation, in this work, following the 
decoupled planning paradigm~\cite{ErdLoz86}, we perform a systematic study of an 
intuitive principle for the design of better performing heuristics for \mpp. The 
decoupled setting generally involves two planning phases, where the first phase 
plans individual robot paths ignoring other robots and the second one resolves 
robot-robot conflicts within some spatio-temporal window. Traditionally, this 
phase is executed by running single robot path plannings ignoring other robots. 
The \emph{Space Utilization Optimization} (\suo) principle tackles the first 
planning phase, seeking to make robots use the free space ``evenly''. 

Based on vertex, edge, and temporal usage information, \suoi, as our 
implementation of the \suo principle, builds a global heuristic that tracks 
how the free space is being used among all participating robots. 
We then exploit applying \suoi as both an \emph{estimated cost-to-go}, 
for reducing congestion, and as part of the \emph{cost-to-come}, for 
reducing conflicts along the entire robot path. Theoretically, we prove 
that \suoi does not compromise individual path optimality while simultaneously 
achieves its design goal of providing better space utilization. In practice, 
\suoi leads to significant improvement in initial path quality, resulting in 
over $40\%$ reduction of path conflicts. 


The introduction of \suoi brings notable improvement to a multitude of \mpp 
and \mppl benchmarks. \suoi, which may be applied as an orthogonal heuristic to 
many \mpp algorithms, leads to sizable gains in both computation time and 
solution optimality when combined with efficient methods for the second phase
of a decouple planner for path scheduling, like DDM~\cite{han2020ddm} and 
ECBS~\cite{barer2014suboptimal}. For example, using \suoi with the database-driven 
collision resolution from \cite{han2020ddm} leads to $15\%+$ reduction in 
computation time and improves solution optimality by roughly $25\%$. 
For \mppl, \suoi, with additional planning horizon management, could reduce the 
computation time of a state-of-the-art methods \cite{li2020lifelong} by more than 
$65\%$, while keeping the same level of optimality.

\textbf{Related Work. }
Multi-Robot Path Planning (\mpp), or equivalently, Multi-Agent Path Finding 
(\mapf)~\cite{stern2019multi}, has been actively studied for decades from many 
angles including computational complexity and effective algorithm 
design~\cite{Gol84, ErdLoz86, GuoPar02}. 
%
Until a few years back, studies on \mpp focus mainly on \emph{one-shot} or 
\emph{static} problems, where $n$ robots are to reach $n$ specific goals. 
Many algorithms for computing high-quality or optimal solutions have been proposed.
Decoupled solutions~\cite{ErdLoz86} dominate the algorithmic attack, with methods 
using techniques including independence detection~\cite{StaKor11}, sub-dimensional 
expansion~\cite{wagner2015subdimensional}, 
conflict-based search~\cite{boyarski2015icbs, cohen2016improved}, among others. 
Methods have also been proposed through the reduction to other problems including
satisfiability (SAT)~\cite{Sur12}, 
Answer Set Programming (ASP)~\cite{erdem2013general}, 
and multi-commodity flow~\cite{YuLav16TRO}. 
There also exists prioritized 
methods~\cite{BerOve05, bennewitz2002finding, saha2006multi,BerSnoLinMan09} 
and a divide-and-conquer approach~\cite{yu2018constant} 
which achieve good scalability but at the cost of either completeness or optimality. 
In \cite{vedder2019x}, an any time algorithm is proposed to quickly find a feasible 
solution, which is subsequently improved. 
A learning-assisted approach~\cite{sigurdson2019automatic} has been developed to 
automatically select the algorithm for solving \mpp challenges.

With the rise of multi-robot applications in the logistics domain~\cite{WurDanMou08},
\emph{dynamic} or \emph{life-long} \mpp variants, or \mppl, have attracted 
attentions in the past few years. Recent work has focused on dynamic warehouse 
setups, pursuing both better planning algorithms~\cite{ma2017lifelong} and robust 
execution schedules~\cite{hoenig2019persistent}. Prioritized planning method 
with a flexible priority sequence has also been developed~\cite{okumura2019priority}.

The general idea behind the \suo principle, better usage of the shared free space, 
has been explored under both single and multi-robot settings. For single robot 
exploring a obstacle-laden domain, a path ensemble can increase the chance of 
succeeding in finding a longer horizon 
plan~\cite{branicky2008path, erickson2009survivability}. 
Similar to what we explore in the current study, path diversity is just one of the 
relevant factors in \mpp resolution~\cite{knepper2009path}.
Survivability is also examined under a probabilistic framework for multi-robot 
systems \cite{lyu2016k}. Under a similar context, a heuristic based on path conflicts 
expedites the solution process of an \mpp algorithm~\cite{barer2014suboptimal}.

Despite the fact that the \suo principle is intuitive, to our knowledge, our 
exploitation of the principle, building on our initial work \cite{han2020ddm}, 
makes novel contributions to the field. In contrast to \cite{han2020ddm}, this study 
{\em (i)} thoroughly exploits the \suo principle with the introduction of the \suoi 
heuristics with a number of variations;
{\em (ii)} proves \suoi's collision avoidance property: 
it finds the shortest paths while minimizes certain collision-based metrics; 
{\em (iii)} integrates \suo to \mpp and \mppl algorithms and empirically shows that 
using \suoi  benefits both computation time and solution optimality.




\section{Preliminaries}
\subsection{Problem Statement}
Multi-Robot Path Planning (\mpp) tasks to find collision-free paths 
that efficiently route robots. 
Consider an undirected graph $\mathcal G(V, E)$ and $n$ robots 
with start configuration $S = \{s_1, \dots, s_n\} \subseteq V$ and goal configuration 
$G = \{g_1, \dots, g_n\} \subseteq V$. Each robot has start and goal vertices $s_i$, $g_i$.
We define a {\em path} for robot $i$ as a map 
$P_i: \mathbb {N} \to V$ where $\mathbb N$ is the set of non-negative integers. 
A feasible $P_i$ must be a sequence of vertices that connects $s_i$ and $g_i$: 
1) $P_i(0) = s_i$;
2) $\exists T_i \in \mathbb N$, s.t. $\forall t \geq T_i, P_i(t) = g_i$;
3) $\forall t > 0$, $P_i(t) = P_i(t - 1)$ or $(P_i(t), P_i(t - 1)) \in E$. 

Here, we first define the single robot version of the problem. 

\begin{problem}{\bf Single-Robot Path Planning (\spp).}\label{prob:spp}
    Given $\mathcal G, s, g$, find a feasible path $P$.
\end{problem}

Following the feasibility definition of $P_i$, we denote $T_i$ as the 
{\em length} of $P_i$. 
We call $P_i$ the {\em shortest path} for robot $i$ if and only if it minimizes $T_i$. 
Given a set of paths $\{P_1, \dots, P_n\}$ for all robots, 
we call them {\em collision-free} if and only if there are no simultaneous 
occupancy of the same vertex or edge. 
That is, $\forall 1 \leq i < j \leq n$, $P_i, P_j$ must satisfy: 
1) $\forall t \geq 0$, $P_i(t) \neq P_j(t)$;
2) $\forall t > 0$, $(P_i(t - 1), P_i(t)) \neq (P_j(t), P_j(t - 1))$. 

We say paths are {\em independent} if they are 
feasible single robot paths but not necessarily collision-free.

The traditional one-shot \mpp problem is defined as

\begin{problem}{\bf Multi-Robot Path Planning (\mpp).}\label{prob:mpp}
    Given $\mathcal G, S, G$, find a collision-free path set $\{P_1, \dots, P_n\}$.
\end{problem}

An {\em optimal} solution for \mpp may minimize the {\em makespan} 
$\max_{1 \leq i \leq n} |T_i|$ 
or {\em sum-of-cost} $\sum_{1 \leq i \leq n} T_i$. 

Apart from \mpp, we also study the {\em life-long} 
variation \mppl where each robot has {\em a list of goal vertices}. 
We denote the goal configuration as $\mathbf G = \{\mathbf g_1, \dots, \mathbf g_n\}$ 
where $\mathbf g_i = (g_i^1, g_i^2, \dots)$. 
Here, $g_i^k$ is the $k$-th goal in robot $i$'s goal list $\mathbf g_i$.
Note that  for an actual \mppl instance, $\mathbf g_i$ is often a list that 
is constantly updated. 
In \mppl, the second condition for a feasible path $P_i$ becomes: 
2) $\exists T_i^1 < T_i^2 < \dots \in \mathbb N$, s.t. $P_i(T_i^k) = g_i^k$. 

With all other conditions inherited from \mpp, we have 

\begin{problem}{\bf Life-long Multi-Robot Path Planning (\mppl).}\label{prob:mppl}
    Given $\mathcal G, S, \mathbf G$, find a collision-free path set 
    $\{P_1, \dots, P_n\}$.
\end{problem}

\mppl algorithms often optimizes {\em throughput}, i.e., the average 
number of goal reaches in a unit time step. 
Given a large $T \in \mathbb N$, the throughput can be expressed by 
$(\sum_{1 \leq i \leq n}\argmax_{k}  (T_i^k | T_i^k \leq T)) / T$. 

For practicality and simplifying explanation, we assume $\mathcal G$ is a 
$4$-connected grid. All algorithms and heuristics proposed in this paper 
apply to arbitrary graphs.

\subsection{Importance of \spp in Solving \mpp and \mppl}\label{subsec:spp-mpp}

Solving \spp is a stepping stone toward \mpp and \mppl. 
Many \mpp solvers incorporate \spp planners as sub-routines, e.g., 
decoupled planners generally use a two-phase approach to first plan 
independent paths and then resolve collisions.
Such methodologies are popular for two reasons.
First, unlike multi-robot planning which is hard to 
optimize \cite{Yu2015IntractabilityPlanar}, 
\spp has been thoroughly studied and can be solved efficiently and optimally 
using A* search with simple heuristics. 
Second, in a practical setting with relatively low robot density, 
usually only small sub-groups of robots have local interactions 
at any given time, which can be quickly resolved.

\begin{wrapfigure}[6]{r}{0.25\linewidth}
    \hspace*{-10pt}
    \includegraphics[width = \linewidth]{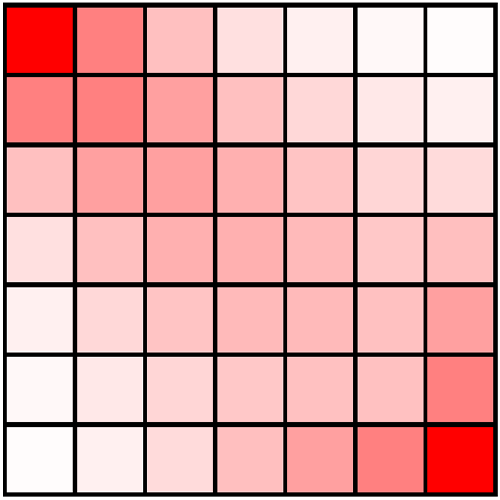}
\end{wrapfigure}
However, such a two-phase approach tends to overuse parts of the free space while leaving 
other parts underutilized. To demonstrate the effect, we plan shortest paths 
from the top left corner to the bottom right corner on a $7 \times 7$ 
grid with randomized node exploration and randomized tie-breaking. 
The shades of cells in the figure visualizes the probability a 
cell is to be used. The paths concentrate along the diagonal connecting 
the two corners, leaving the top right and bottom left corners underutilized. 
This effect gets more pronounced when obstacles exist and the available 
path choices are limited. 

\mppl, more general and practical than \mpp, is often solved iteratively 
using existing \mpp algorithms. Thus, unbalanced graph utilization issue also 
negatively impacts \mppl solvers. 

\section{\suo: Principles and Implementation}\label{sec:suo}
\subsection{Optimizing Space Utilization: Principles}
\begin{figure}[h!]
    \centering
    \subfloat[Without \suoi]{\includegraphics[width=0.32\linewidth]{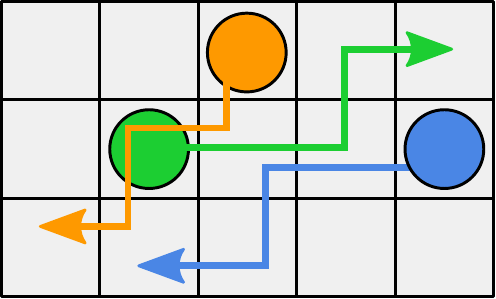}\label{fig:example-wo-suo}}
    \hspace{20pt}
    \subfloat[With \suoi]{\includegraphics[width=0.32\linewidth]{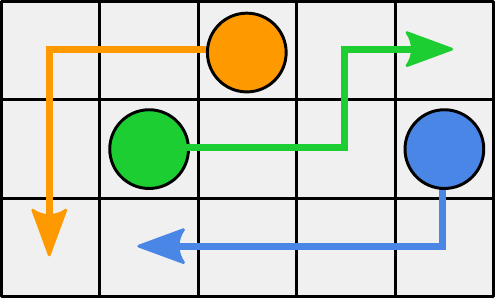}\label{example-w-suo}}
    \caption{Two sets of individual paths in a $5 \times 3$ grid. 
    The robots' start configuration is shown as colored disks. 
    The planned individual paths are drawn as colored lines. 
    (a) Randomly generated paths may result in congestion. 
    (b) When using \suoi, which ``spreads out'' the robots' paths, 
    the number of conflicts is reduced.
    }
    \label{fig:example-suo}
\end{figure}

From the discussion in Section~\ref{subsec:spp-mpp}, robots' individual paths 
should be spread ``evenly'' across the free space to reduce congestion.  
We call this idea the {\em Space Utilization Optimization (\suo) 
principle}. In this work, we develop {\em a first \suo implementation}, 
\suoi (SU-First), which 
serves as a global heuristic to help generate independent paths for 
\mpp/\mppl algorithms to better utilize graph resources. 
The example in Fig.~\ref{fig:example-suo} demonstrates that \suoi 
could potentially 
enhance both computational efficiency and optimality. 


In \suoi, each path uses vertex, edge, and temporal information from other paths 
to avoid more congested areas or to reduce the sum of conflicts 
along the path. 
\suoi improves upon its predecessor~\cite{han2020ddm} which only considered vertex 
information.
In Appendix~\ref{app:examples}, we use several examples to highlight the essence of 
vertex, edge, and temporal information when applying \suo. 


\subsection{Path Planning with \suoi: High Level Procedure}

We now describe the implementation of \suoi, including 
theoretical guarantees and evaluation results to show 
that \suoi with the proposed path planning procedure indeed optimizes 
space utilization. Since \mppl algorithms are often based on \mpp solvers,
we only describe \suoi in the context of \mpp; \suoi is directly
applicable to \mppl.

\begin{algorithm}
\begin{small}
\DontPrintSemicolon
\caption{\label{alg:search} Generate independent paths with \suoi}
\KwIn{Graph $\mathcal G$, $n$ robots with starts $S$ and goals $G$}
$\pi \gets \textsc{GetOrderByDistanceToGoal}(\mathcal G, S, G)$\;\label{alg:search:order}
$S \gets \pi(S), G \gets \pi(G)$\;\label{alg:search:sort}
\lFor{$1 \leq i \leq n$}{$P_i \gets {\text{None}}$}
\For{number of planning iterations $r$}{\label{alg:search:multiple}
\For{$1 \leq i \leq n$}{\label{alg:search:for}
    $\mathcal T \gets \textsc{BuildSuo}(P_1, \dots, P_{i-1}, P_{i+1},\dots, P_n)$\;\label{alg:search:suo}
    $P_i \gets \textsc{FindPath}(\mathcal G, s_i, g_i, \mathcal T)$\;\label{alg:search:search}
} 
}
\Return{$\pi^{-1}(P_1, \dots, P_n)$}
\end{small}
\end{algorithm}

Alg.~\ref{alg:search} shows the high level procedure that uses \suoi 
to generate individual paths. 
In lines~\ref{alg:search:order}-\ref{alg:search:sort}, robots are sorted in 
{\em descending} distance-to-goal order. 
Then, paths for the robots are planned sequentially 
(lines~\ref{alg:search:for}-\ref{alg:search:search}) while using \suoi to 
avoid previously planned paths. 
The rationale for the descending ordering is that planning longer paths first helps  
quickly collect graph utilization information. 
%
As indicated in line~\ref{alg:search:multiple}, the path planning procedure 
is repeated for $r$ iterations as more iterations capture the space 
utilization more accurately, improving the collision avoidance. 
Both robot ordering and number of iterations are tested in Section~\ref{sec:evaluation}.

\subsection{\suoi Heuristic Construction and Computation}
At line~\ref{alg:search:suo}, before each time 
we call \spp planner, \textsc{BuildSuo} first builds an \suoi lookup table $\mathcal T$
to memorize the space usage of existing paths. 
Denoting $\mathcal T(v, t)$ (resp. $\mathcal T(v_1, v_2, t)$) as the expected 
usage of vertex $v$ (resp. directed edge $v_1, v_2$) at time step $t$, 
$\mathcal T$ aggregates vertex, 
edge, and temporal information over all \textsc{BuildSuo} input paths: 
\begin{align*}
    \mathcal T (v, t) & = \textstyle \sum_{P} [(v) \preceq P_i], 
\end{align*}
\begin{align*}
    \mathcal T (v_1, v_2, t) & = \textstyle \sum_{P} [(v_1, v_2) \preceq P_i].
\end{align*}
Here, $[\cdot]$ on the right hand side of the equations is an \emph{indicator} variable: 
it is $1$ if the expression inside is true, otherwise it takes $0$. 
We use $\preceq$ to denote a sub-sequence relationship.


\suoi is mainly used to generate initial paths for \mpp algorithms. 
Since the initial paths can be modified (e.g. delayed or diverted 
for collision avoidance) in a full \mpp algorithm, 
we add integer parameters $\alpha_L \geq 0$ and $\alpha_H \geq 0$ 
to reason about graph usage in adjacent time steps 
and handle robot movement uncertainty: 
\begin{align*}
    \mathcal T (v, t) & = \textstyle \sum_{P} [(v) \preceq P(t - \alpha_H: t + \alpha_L)], 
\end{align*}
\begin{align*}
    \mathcal T (v_1, v_2, t) & = \textstyle \sum_{P} [(v_1, v_2) \preceq P(t - \alpha_H: t + \alpha_L)].
\end{align*}
For each path, the occupancy of graph utility at time $t$ 
can now expand its influence to time steps in $(t - \alpha_L, t + \alpha_H)$. 
The values of $\alpha_L, \alpha_H$ are empirically determined based on 
the \mpp algorithm itself; in Section~\ref{sec:evaluation}, we observe that such an 
temporal reasoning feature delivers better collision avoidance results when 
the movement of robots is uncertain.  

We use $\mathcal T$ to calculate \suoi heuristic and then use it during 
\spp path planning (line~\ref{alg:search:search}). 
Given a state transition $(v_1, v_2, t)$ which means the robot moves 
from vertex $v_1$ to vertex $v_2$ at time $t$, the \suoi heuristic value is 
\begin{equation*}\label{equation:suo}
    H_\text{\suoi}(v_1, v_2, t) = \beta_v \frac{\mathcal T(v_2, t)}{n} + 
    \beta_e \frac{\mathcal T(v_2, v_1, t)}{n}.
\end{equation*}
Here, the first term indicates the amount of vertex conflicts the robot 
may encounter for the inferred state transition, while the second term 
is associated with head-to-head edge collisions. 
Parameters $\beta_v, \beta_e>0$ are used to balance between vertex and 
edge information; $\beta_v + \beta_e = 1$.
When each vertex/edge in the graph is visited for at most once by each single robot path, we have
\begin{lemma}\label{lemma:occupancy-bound}
When $\beta_v + \beta_e = 1$, $0 \leq H_\text{\suoi}(\cdot) < 1$. 
\end{lemma}

The condition $H_\text{\suoi}(\cdot) < 1$ is essential for 
\suoi to behave as a tie-breaker, 
which facilitates a good balance between single path optimality and congestion avoidance.  

\begin{remark}\label{remark:no-temp}
A special case for constructing $\mathcal T$ is to ignore temporal information, 
i.e., instead of using $(v, t)$ and $(v_1, v_2, t)$ as lookup table keys, we 
use $(v)$ and $(v_1, v_2)$. Thus, $\mathcal T(v)$ (resp., $\mathcal T(v_1, v_2)$) simply 
records the total number of times the vertex $v$ (resp., the edge $(v_1, v_2)$) 
is used by the existing paths. This leads to smaller lookup tables 
but potentially worse collision avoidance as a result. 
\end{remark}

\begin{remark}
Note that for the actual implementation, in Alg.~\ref{alg:search}, line~\ref{alg:search:suo}, 
$\mathcal T$ is not re-constructed but updated based on the previous iteration which makes 
the computational complexity for the construction step $O(|P|(\alpha_L + \alpha_H))$.
\end{remark}

Line~\ref{alg:search:search} uses standard A* to find a path from $s_i$ to $g_i$. 
In the next two subsections, we discuss two ways to integrate \suoi into A*. 
For simplicity, we now assume \suoi only uses vertex information 
(i.e., $\beta_v = 1, \beta_e = 0$) without temporal information 
(see Remark~\ref{remark:no-temp}), unless otherwise specified. 


\subsection{\suoi as Part of Estimated Cost-To-Go}\label{sec:suo-cost-to-go}
We use $H_{\text{short}}(v, g_i)$ to denote the shortest path 
distance between $v$ and $g_i$ in $\mathcal G$. 
$H_{\text{short}}(v, g_i)$ is graph-dependent and can be calculated before 
path planning. For a grid without obstacles, $H_{\text{short}}(v, g_i)$ is 
the Manhattan distance heuristic. The heuristic we use in A* search is 
\[\mathcal H(v) = H_{\text{short}}(v, g_i) + H_{\suoi}(v).\]


\begin{lemma}\label{lemma:shortest}
A path planned using A* search with $\mathcal H$ as heuristic 
is a shortest path from $s_i$ to $g_i$.
\end{lemma}
\vspace*{-6pt}
\begin{proof}
See Appendix~\ref{app:proof}.
\end{proof}
\vspace*{-3pt}

$\mathcal H$ not only ensures a shortest path is found; the path also
\emph{minimizes the maximum single step conflict}. 
Given path $P_i$, its maximum single step 
conflict can be represented as 
\[\textstyle C_{\text{single}}(P_i) = \max_{0 < t < |P_i|} \mathcal T[P_i(t)].\]

\begin{lemma}\label{lemma:minimum}
A path $P_i$ planned using A* search with $\mathcal H$ heuristic 
is a shortest path minimizing {\normalfont $C_{\text{single}}(P_i)$}.
\end{lemma}
\vspace*{-6pt}
\begin{proof}
See Appendix~\ref{app:proof}.
\end{proof}
\vspace*{-3pt}

Now, given paths $P_1, \dots, P_n$ returned from Alg.~\ref{alg:search}, 
denoting the maximum conflict on a single vertex as 
\[\textstyle \mathcal C_{\text{single}} = 
\max_{v \in \mathcal G} \sum_{1 \leq i \leq n} [v \in Im(P_i)],\]
we reach the following property. 

\begin{lemma}\label{lemma:cost-to-go-converge}
$\mathcal C_{\text{single}}$ cannot increase after the first \suoi planning iteration. 
\end{lemma}
\vspace*{-6pt}
\begin{proof}
See Appendix~\ref{app:proof}.
\end{proof}
\vspace*{-3pt}

Lemma~\ref{lemma:cost-to-go-converge} directly leads to the following theorem. 
\begin{theorem}
$\mathcal C_{\text{single}}$ will converge as the number of planning iterations increases. 
\end{theorem}

\subsection{\suoi as Part of Cost-To-Come}

With the default transition cost as $1$ for all states during A* search, 
when aggregating \suoi it into cost-to-come. 
we define the new transition cost leading to vertex $v$ as 
\begin{equation*}\label{equation:cost}
    \textstyle
    C(v) = 1 + H_{\suoi}(v) / (\max_{1 \leq i \leq n} H_\text{short}(s_i, g_i) + 1) .
\end{equation*}
The collision avoidance property of using \suoi with cost-to-come is different from 
that of using \suoi with cost-to-go, 
We hereby define the number of vertex collisions on $P_i$ as 
\begin{align*}
    C_\text{path}(P_i)
    &= \textstyle \sum_{1 \leq j \leq n, j \neq i} |\text{Im}(P_i) \cap \text{Im}(P_j)|. 
\end{align*}

\begin{lemma}\label{lemma:cost-to-come-shortest}
A path $P_i$ planned using A* search with $C$ as transition cost 
and an admissible heuristic 
is the shortest path which minimizes {\normalfont $C_{\text{path}}(P_i)$}.
\end{lemma}
\vspace*{-6pt}
\begin{proof}
See Appendix~\ref{app:proof}.
\end{proof}
\vspace*{-3pt}

Given resulting paths as $P_1, \dots, P_n$ when using \suoi as cost-to-come, 
we denote the total number of collisions as 
\[\textstyle \mathcal C_{\text{path}} = \sum_{1 \leq i \leq n} C_{\text{path}}(P_i), \]
we find the following property. 

\begin{lemma}\label{lemma:cost-to-come-converge}
$\mathcal C_{\text{path}}$ cannot increase after the first \suoi iteration. 
\end{lemma}
\vspace*{-6pt}
\begin{proof}
See Appendix~\ref{app:proof}.
\end{proof}
\vspace*{-3pt}

Lemma~\ref{lemma:cost-to-come-converge} directly leads to the following theorem. 
\begin{theorem}
$\mathcal C_{\text{path}}$ will converge as the number of planning iterations increases. 
\end{theorem}

\begin{remark}
Regardless of whether \suoi is used as part of the cost-to-come or the cost-to-go, 
even though the properties mentioned above are only proved for \suoi without 
temporal information, similar properties exist for \suoi with temporal information 
when using state-time A*. 
Instead of just considering vertex conflicts, 
all lemmas in this section remain true when edge conflicts are considered (i.e. $\beta_e > 0$), 
except for Lemma~\ref{lemma:cost-to-go-converge} and Lemma~\ref{lemma:cost-to-come-converge}. 
However, in evaluation, we empirically observe that running multiple planning iterations 
considering edge conflicts is still beneficial. 
\end{remark}



\section{Space Utilization Optimization Application}

The paths generated by \suoi can be directly used as input to some \mpp/\mppl 
algorithms. 
The more balanced graph utilization and reduced conflicts facilitate collision 
resolution, improving both computation time and solution optimality. 

\suoi can also be combined with time-based divide-and-conquer to provide better intermediate goals. 
With a baseline structure adapted from \cite{li2020lifelong}, 
we first propose a {\em horizon cut} technique to reduce unnecessary 
node explorations and then use \suoi to further enhance the performance. 

\subsection{Baseline Bounded-Horizon Search for \mppl}\label{sec:algorithm:lifelong}

It is well known that solving an entire \mppl instance in one-shot is impractical, 
not only because the long lists of goals makes the problem computationally demanding,  
but also because the goal lists could be dynamically updated in real world scenarios, 
which invalidates the current solution and brings the need for online re-planning. 

Given the above factors, \mppl is usually solved by a bounded-horizon approach, 
The basic idea is to plan the paths for the $h$ time steps, 
execute the paths, and then iteratively re-plan and execute. 
Here, $h$ is called the {\em planning horizon}. 
The pseudocode of such a baseline horizon-based structure~\cite{li2020lifelong} is 
provided in Alg.~\ref{alg:horizon}; readers may ignore 
lines~\ref{alg:horizon:cut-start}-\ref{alg:horizon:cut} for now as we will 
discuss later. 
In the beginning, the goal list $\mathbf{g}_i$ for each robot $i$ is shortened 
until there is at most one unreachable goal for horizon $h$ 
(lines~\ref{alg:horizon:truncate:1}-\ref{alg:horizon:truncate:2});
we use $(-1)$ to index the last element in a sequence, and $+$ to indicate sequence 
concatenation. 
Then, in line~\ref{alg:horizon:search}, the current state (denoted as $S$) 
and the modified goal list are sent to a windowed \mpp solver. 
The behavior of the windowed solver is to output an $h$-step collision-free path set, 
where each robot $i$ aims to traverse the shortened goal list $\mathbf{\hat{g}}_i$ in order. 
The selection of the windowed \mpp solver is flexible. 
In this work, we use Bounded-Horizon Enhanced Collision Based Search~\cite{li2020lifelong} 
with weight parameter $w = 1.5$ and treat it as a black box. 

\begin{algorithm}
\small
\DontPrintSemicolon
\caption{\label{alg:horizon} \textsc{BoundedHorizonSearch}}
\KwIn{Graph $\mathcal G$, current state $S$, goal lists $\mathbf G$, horizon $h$.}
\KwOut{A $h$-step solution}
$\forall 1 \leq i \leq n$, $\mathbf{\hat{g}}_i \gets (s_i)$, $d_i \gets 0$\;\label{alg:horizon:truncate:1}
\For{$1 \leq i \leq n$}{
    \For{$g \in \mathbf{g}_i$}{
        $\mathbf{\hat{g}}_i \gets \mathbf{\hat{g}}_i + (g)$\;
        $d_i \gets d_i + \textsc{ShortestDistance}(\mathbf{\hat{g}}_i(-1), g)$\;
        \lIf{$d_i \geq h$}{break\label{alg:horizon:truncate:2}}
    }
}
\If{using horizon cut}{\label{alg:horizon:cut-start}
$\forall 1 \leq i \leq n$, $P_i \gets ()$\;
\For{$1 \leq i \leq n$}{
    \If{using \suoi}{      
        $H_\suoi \gets \textsc{BuildSuo}(P_1, \dots, P_n)$\;\label{alg:horizon:suo2}
        $P_i \gets \textsc{FindPath}(\mathcal G, \mathbf{\hat{g}}_i(-2), \mathbf{\hat{g}}_i(-1), H_\suoi)$\label{alg:horizon:suo1}
    }
    \lElse{
        $P_i \gets \textsc{FindPath}(\mathcal G, \mathbf{\hat{g}}_i(-2), \mathbf{\hat{g}}_i(-1))$\label{alg:horizon:simple-cut}
    }
    $\mathbf{\hat{g}}_i(-1) \gets P_i(- (|P_i| - d_i + h - 1))$\;\label{alg:horizon:cut}
}
}
\Return{$\textsc{WindowedSolver}(S, \mathbf{\hat{g}}_1(1:), \dots, \mathbf{\hat{g}}_n(1:), h)$}\label{alg:horizon:search}
\end{algorithm}


\subsection{Horizon Cut and \suoi Integration}

\begin{figure}[h!]
    \centering
    \subfloat[Baseline]{\includegraphics[width=0.32\linewidth]{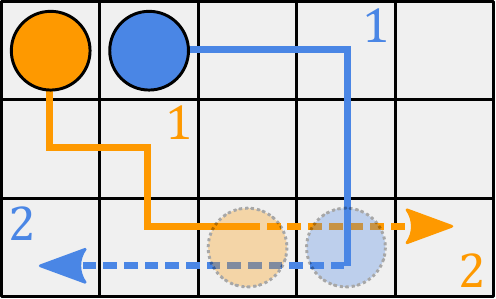}\label{fig:horizon-comparison:baseline}}
    \hspace*{1pt}
    \subfloat[Horizon cut]{\includegraphics[width=0.32\linewidth]{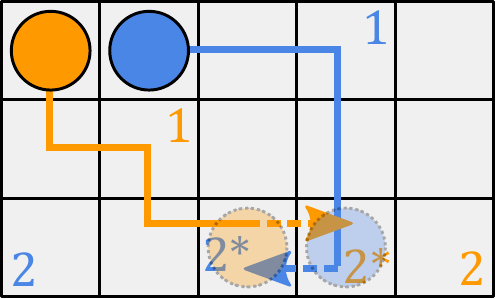}\label{fig:horizon-comparison:cut}}
    \hspace*{1pt}
    \subfloat[Adding \suoi]{\includegraphics[width=0.32\linewidth]{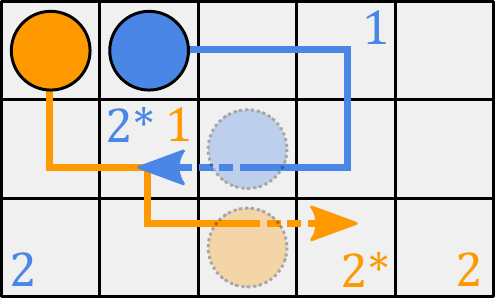}\label{fig:horizon-comparison:suo}}
    \caption{An example comparing bounded-horizon baseline, horizon cut, 
    and horizon cut with \suoi. 
    The robots each has $2$ goals to reach, marked as colored numbers. 
    The planned configuration after the current horizon ($h = 4$) is visualized as transparent disks.
    (a) The paths planned by the baseline bounded-horizon method. 
    The dashed part is planned but not executed. 
    (b) Using horizon cut, we avoid planning unnecessary steps by setting the last goals as $2^*$. 
    (c) With \suoi, we have a better selection of $2^*$ so that the conflicts in 
    the next planning horizon may be avoided beforehand. 
    }
    \label{fig:horizon-comparison}
\end{figure}

Due to using \spp algorithm as subroutine, the baseline bounded-horizon 
method wastes computation power due to unnecessary reasoning about 
paths outside horizon $h$: although these paths are not collision-checked, 
they are planned by the windowed solver and are discarded afterwards. 
To overcome this weakness, we propose {\em horizon cut}, 
which reduces the number of search nodes generated outside of $h$ 
while still ensures that the robots move toward their future goals. 
Horizon cut further truncate the goal list by changing the last goal to a vertex that is 
in-between the second last goal (i.e., $\mathbf{\hat{g}}_i(-2)$) and the last goal 
(see Alg.~\ref{alg:horizon}, lines~\ref{alg:horizon:simple-cut}-\ref{alg:horizon:cut}). 
In our implementation, we find a shortest path between $\mathbf{\hat{g}}_i(-2)$ and 
$\mathbf{\hat{g}}_i(-1)$ and select the vertex at $t = h + 1$. 
The dashed paths in 
in Fig.~\ref{fig:horizon-comparison:baseline} and Fig.~\ref{fig:horizon-comparison:cut} 
demonstrate that we can avoid planning redundant paths when using horizon cut. 

We then integrate \suoi to help select better target vertices 
(see Alg.~\ref{alg:horizon}, lines~\ref{alg:horizon:suo2}-\ref{alg:horizon:suo1}, 
which is a similar procedure as Alg.~\ref{alg:search}). 
By selecting target vertices on paths which better utilize 
graph resources, we can avoid conflicts in the future planning iterations. 
We visualize the effect in Fig.~\ref{fig:horizon-comparison:suo}, 
where we anticipate less conflicts between the two robots in 
the next planning and execution iteration when using \suoi. 

\section{Evaluation}\label{sec:evaluation}

\begin{figure}[h!]
    \centering
    \subfloat[$30 \times 20$ random graph]{\includegraphics[height=0.22\linewidth]{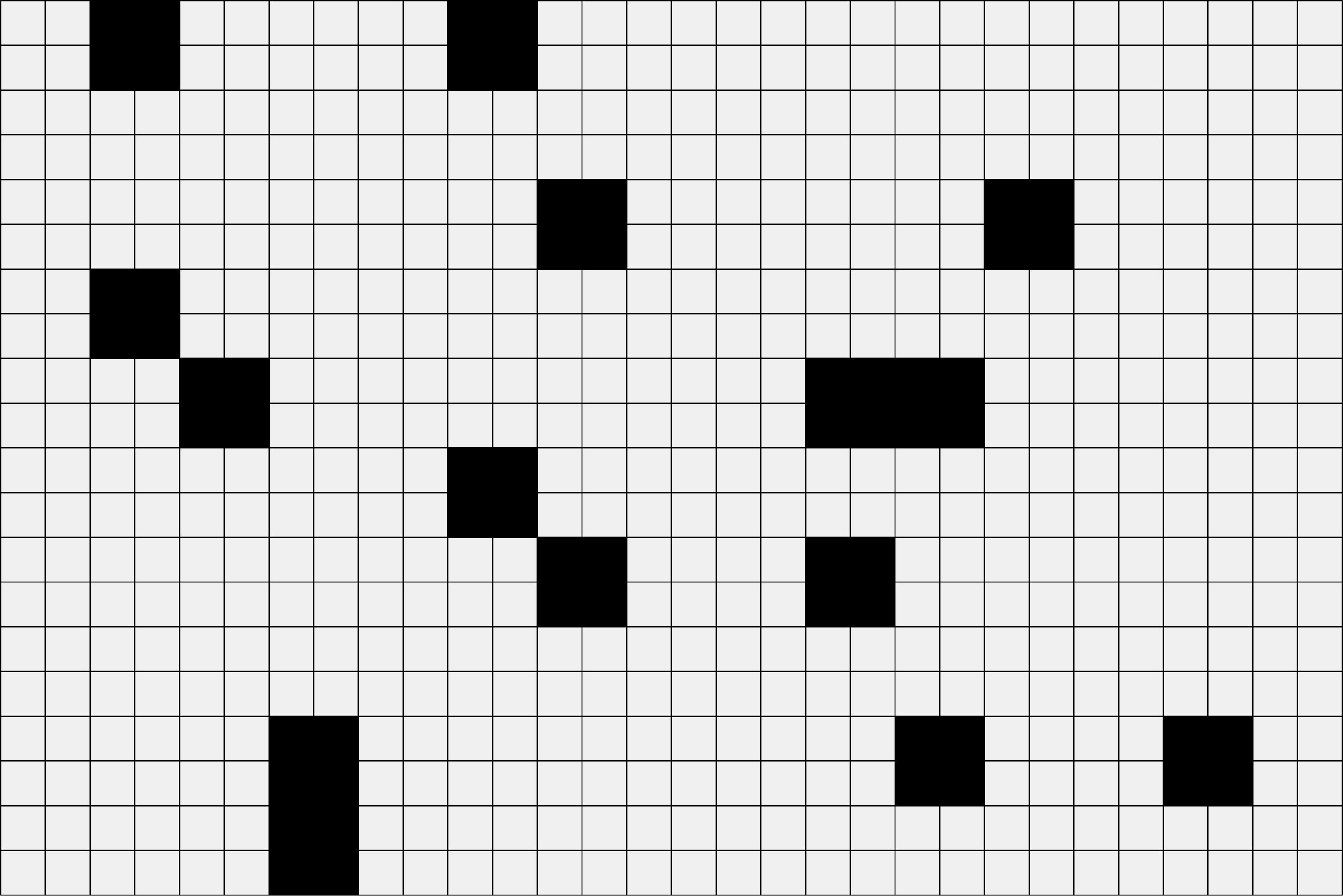}\label{fig:graphs:random}}
    \hspace*{1pt}
    \subfloat[$37 \times 20$ warehouse]{\includegraphics[height=0.22\linewidth]{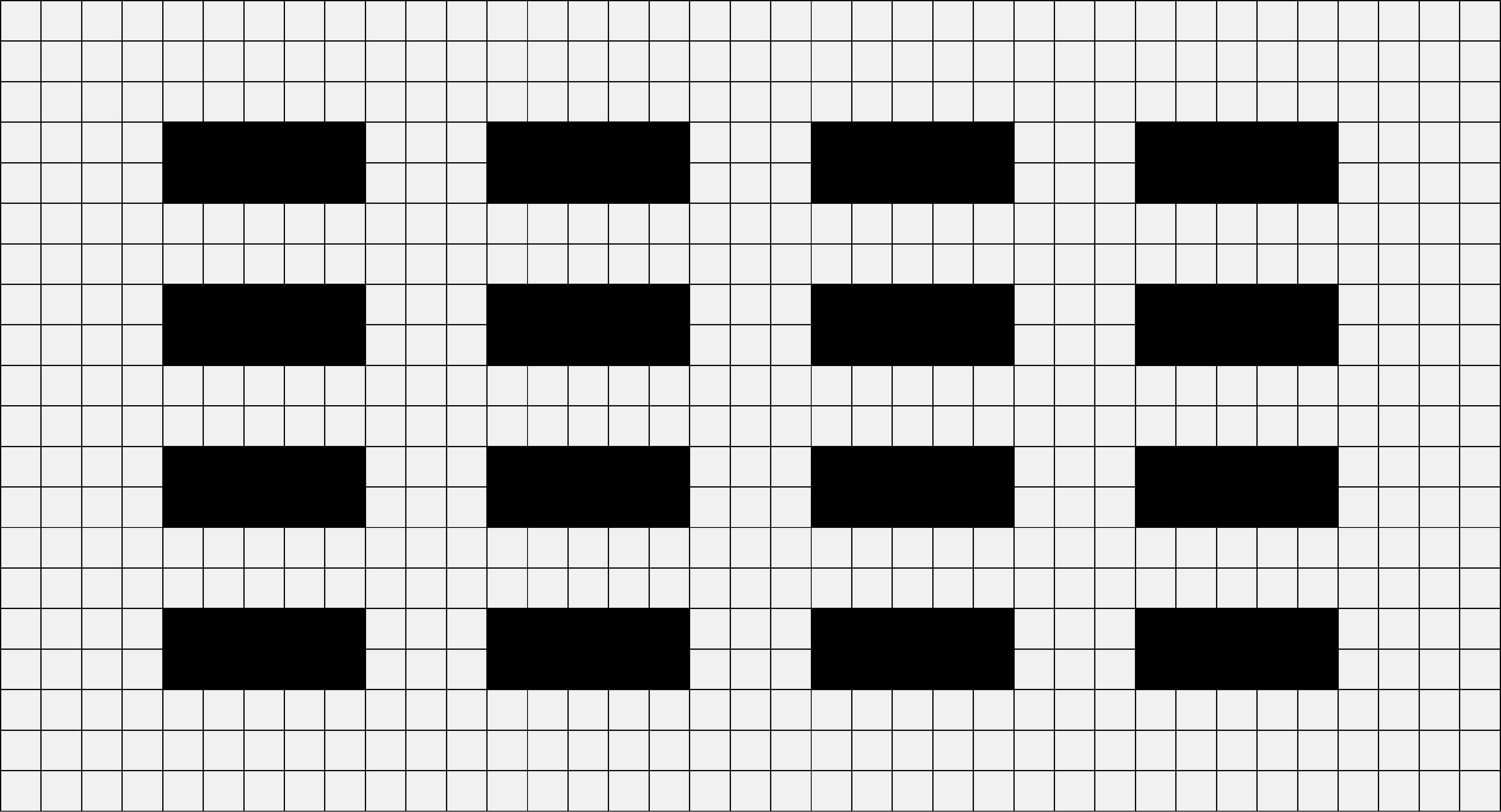}\label{fig:graphs:warehouse}}
    \hspace*{1pt}
    \subfloat[den520d]{\includegraphics[height=0.22\linewidth]{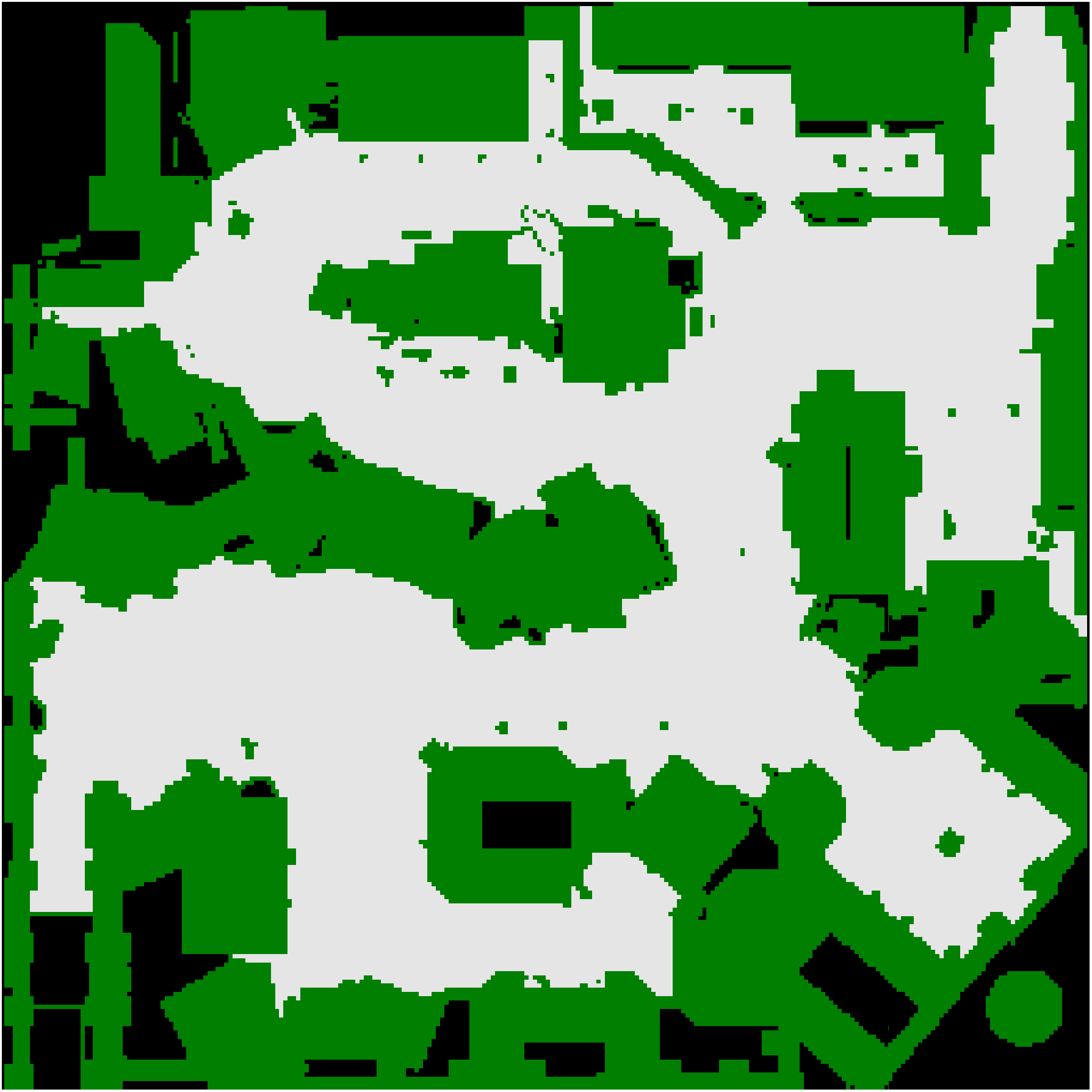}\label{fig:graphs:den520d}}
    \caption{Example of graphs used for evaluation.
    The bright cells visualize vertices. The black and green cells visualize obstacles.
    }
    \label{fig:graphs}
    \vspace{-2mm}
\end{figure}

We performed comprehensive evaluation of \suoi and associated algorithms on 
randomly generated graphs, warehouse-style graphs, and large DAO 
maps (Fig.~\ref{fig:graphs} shows a subset of graphs). All experiments are performed on an 
Intel\textsuperscript{\textregistered} Core\textsuperscript{TM} i7-6900k CPU. 
Data points are averaged over $30$ to $100$ runs on randomly generated problem 
instances. 

Our evaluation focuses on testing \suoi with existing \mpp algorithms. 
Before that, we also conducted analytical experiments, showing that 
the graph utilization conflicts indeed converges to some minimal value with \suoi, 
especially when considering both vertex and edge information with 
a descending ordering of robots. For details, see Appendix~\ref{app:evaluation}. 

\subsection{\suoi in a Full \mpp Algorithm}\label{sec:evaluation:a}

Combining \suoi with an existing collision resolution method makes \mpp solver 
more efficient and optimal. For this evaluation, we used the database-driven 
collision avoidance routine from~\cite{han2020ddm}. The test cases are in $30 
\times 20$ grids with $10\%$ obstacles (see an random example in 
Fig.~\ref{fig:graphs:random}). We report computation time and solution 
optimality (based on sum-of-cost) with varied number of robots. For both 
metrics, lower is better. All data points are normalized in terms of the 
baseline algorithm's performance where \suoi is not used. 

\begin{figure}[h!]
    \centering
    \includegraphics{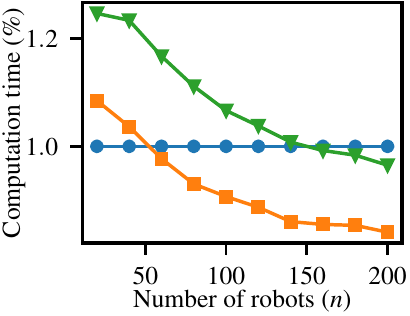}
    \includegraphics{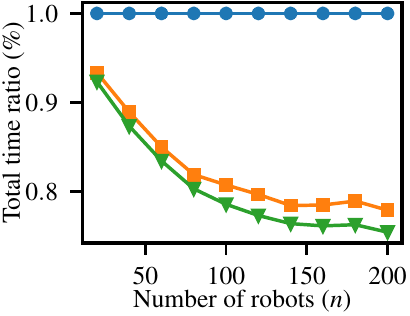}
    \includegraphics{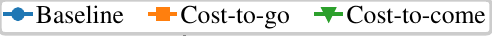}
    \caption{Comparison between using \suoi as a part of estimated cost-to-come 
    and cost-to-go in a full \mpp algorithm.}
    \label{fig:ddm-ctcg}
\end{figure}

Starting with $\beta_v = \beta_e = 0.5$, $r = 1$, and no temporal information, 
we first compare using \suoi as a part of estimated cost-to-go and cost-to-come. 
As shown in Fig.~\ref{fig:ddm-ctcg}, using \suoi as cost-to-go can significantly 
reduce the computation time (by $15\%+$) when robots have interactions. 
As a comparison, using $\suoi$ as cost-to-come only makes the algorithm slightly 
more efficient when the number of robots is large. 
Both \suoi variations significantly improve the optimality, by up to $25\%$. 
The efficiency difference was due to \suoi as cost-to-come minimizes collisions 
along single paths and thus has a larger search space (state-time). 
Hence forth, we use \suoi as part of estimated cost-to-go by default. 

\begin{figure}[h!]
    \centering
    \includegraphics{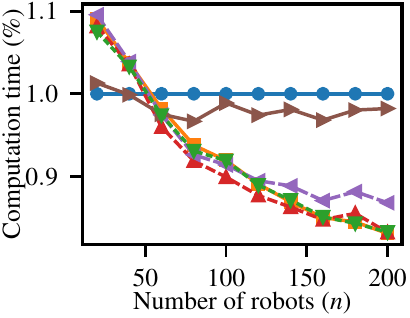}
    \includegraphics{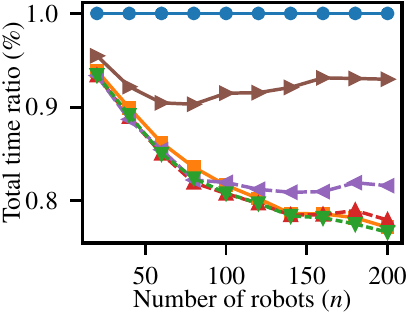}
    \includegraphics{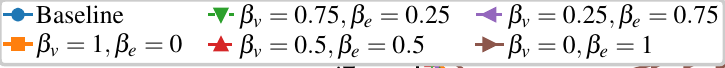}
    \caption{\suoi \mpp solver evaluation with different $\beta_v$, $\beta_e$ values.}
    \label{fig:ddm-ve}
\end{figure}

We then try different $\beta_v$, $\beta_e$ to demonstrate that using vertex and 
edge information together is beneficial. 
Shown in Fig.~\ref{fig:ddm-ve}, the performance improvement of using vertex 
information alone over the baseline is already significant. 
While using edge information alone is weaker than vertex, 
combining the two elements pushes both computation time and solution 
optimality even lower. 
From now on, we set $\beta_v = \beta_e = 0.5$. 

\begin{figure}[h!]
    \centering
    \includegraphics{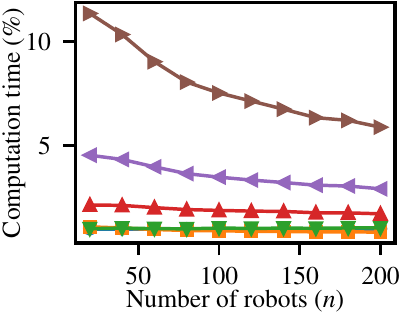}
    \includegraphics{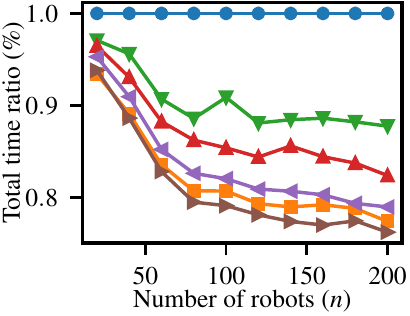}
    \includegraphics{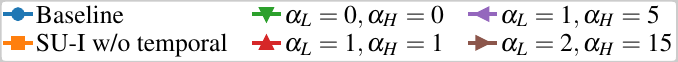}
    \caption{Comparison between different $\alpha_L$, $\alpha_H$ values.}
    \label{fig:ddm-t}
\end{figure}

In Fig.~\ref{fig:ddm-t}, temporal information is added with different $\alpha_L$ 
and $\alpha_H$ values. As stated earlier, due to the modifications to the 
initial paths in the second planning phase, using ``soft'' temporal information 
via extended temporal reasoning is beneficial. This effect is shown here as the 
green line ($\alpha_L = \alpha_H = 0$) under-performs \suoi without temporal 
information, while the brown line ($\alpha_L = 2$, $\alpha_H = 15$) improves 
solution quality. Computation time wise, using temporal information adds overhead 
since it takes a longer time to construct \suoi. So whether to use temporal 
information is a choice to be made by practitioners. 

\begin{figure}[h!]
    \centering
    \includegraphics{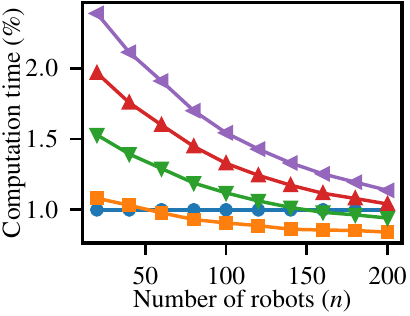}
    \includegraphics{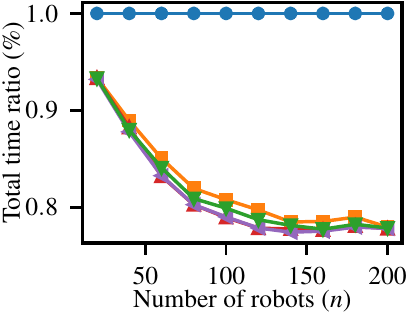}
    \includegraphics{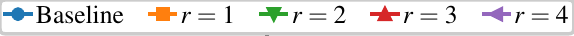}
    \caption{Comparison between different planning iteration $r$.}
    \label{fig:ddm-mr}
\end{figure}

Fig.~\ref{fig:ddm-mr} shows that using multiple planning iterations generates better solutions 
when $r \leq 4$. 

\subsection{\mppl Bounded-Horizon Search with \suoi}

\begin{figure}[h!]
    \centering
    \includegraphics{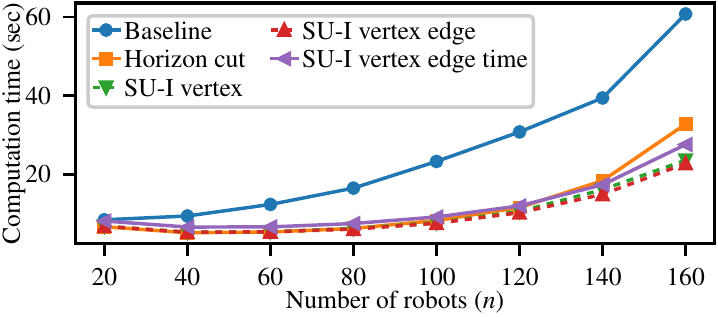}
    \caption{Computation time of bounded-horizon methods.}
    \label{fig:ecbs-time}
\end{figure}

We evaluate \mppl algorithms discussed in Sec.~\ref{sec:algorithm:lifelong} in a 
warehouse-style environment (see Fig.~\ref{fig:graphs:warehouse}). 
Each robot starts from a random vertex and is given a random list of goal vertices. 
The goal lists are continuously extended to make sure each robot always has 
future goals. 
The horizon-based planners with $h = 5$ and internal ECBS~\cite{barer2014suboptimal} 
parameter $w = 1.5$ are called iteratively until a total of $10000$ goals are reached. 
The evaluated methods are the bounded-horizon search~\cite{li2020lifelong} 
(the baseline), with horizon cut, and with \suoi. We report the total computation 
time in Fig.~\ref{fig:ecbs-time}, and the system throughput in Table~\ref{table:ecbs-opt}. 
The throughput is calculated as the average number of goal reached in a single time step; 
the higher, the better. 
The data shows that using horizon cut can effectively decrease the computation time 
by more than $50\%$. At the same time, using \suoi not only makes the computation time 
even lower (by about $65\%$), while keeping a same level of throughput as the baseline. 

\begin{table}[h!]
    \scriptsize
    \setlength\tabcolsep{1mm}
    \centering
    \caption{\label{table:ecbs-opt} Throughput of bounded-horizon methods}
    \begin{tabular}{c|c|c|c|c}
        $n$   & Baseline &Horizon cut &\suoi w/o temporal&\suoi w/ temporal\\ \hline
        $40 $ & 1.981   &1.945        &1.968  &1.974\\ \hline
        $80 $ & 3.737   &3.647        &3.710  &3.719\\ \hline
        $120$ & 5.421   &5.273        &5.370  &5.387\\ \hline
        $160$ & 6.873   &6.714        &6.888  &6.896\\ \hline
    \end{tabular}
\end{table}

\subsection{Bounded-Horizon Search with \suoi on \mpp}\label{sec:evaluation:c}

\begin{figure}[h!]
    \centering
    \includegraphics{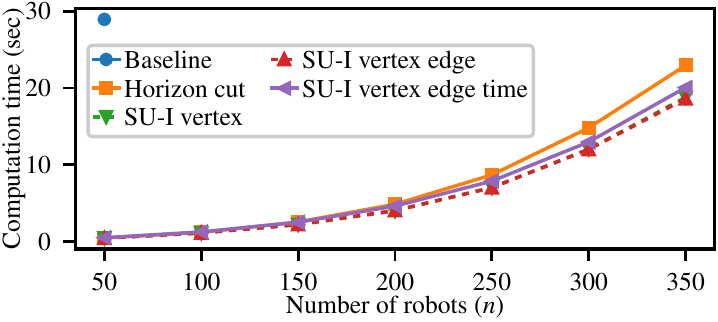}
    \caption{Bounded-horizon search's computation time on a den520d.}
    \label{fig:dao-time}
\end{figure}

As the last evaluation, we directly apply bounded-horizon search to \mpp 
by considering \mpp as a special case for \mppl where all goal lists have length $1$. 
The tested graph is DAO den520d (see Fig.~\ref{fig:graphs:den520d}), a public 
\mpp benchmark~\cite{sturtevant2012benchmarks}. The map size is $257 \times 256$ 
with $28178$ vertices. We set $h = 50$. 
Fig.~\ref{fig:dao-time} shows that \suoi 
remains effective 
in large environments with a sparse robot setup. Note that when not using horizon 
cut and \suoi, the baseline bounded-horizon search is significantly slower due 
to its unnecessary exploration. When we divide the solution quality metric over 
under-estimated lower-bounds (see Table~\ref{table:dao-opt}), we find that our 
method is able to generate solutions very close to the optimal.

\begin{table}[h!]
\scriptsize
    \setlength\tabcolsep{1mm}
    \centering
    \caption{\label{table:dao-opt} Optimality ratio of \mpp solutions in den520d}
    \begin{tabular}{c|c|c|c|c|c|c|c}
        $n$   & 50      & 100     & 150     & 200     & 250     & 300     & 350       \\ \hline
   Makespan   & 1.0002  & 1.0011  & 1.0020  & 1.0031  & 1.0040  & 1.0052  & 1.0064    \\ \hline
Sum-of-time   & 1.0019  & 1.0023  & 1.0034  & 1.0051  & 1.0057  & 1.0076  & 1.0090    \\ \hline
    \end{tabular}
\end{table}

Experiments in Section~\ref{sec:evaluation:a} and Section~\ref{sec:evaluation:c} 
were repeated with different randomly generate graphs, 
more DAO maps (e.g., brc200d, lak201d), and different number of robots. 
The results show similar trends as provided in this section.

\section{Conclusion}
In this work, with \suoi, we performed an in-depth exploration optimizing 
the space utilization for planning better individual robot paths in the first 
phase of a modern decoupled \mpp and \mppl pipeline. In addition to proving 
that \suoi's desirable properties as a heuristic, thorough simulation study 
validates the effectiveness of \suoi in significantly reducing the 
computation load while maintaining or improving the optimality of the resulting 
solution, for both one-shot and life-long multi-robot path planning problems. 

Together with \cite{han2020ddm}, this research opens up a new direction in 
multi-robot path planning. In a sense, \suo, and its implementation, \suoi, are 
taking the decoupled multi-robot path planning paradigm a step further by 
reducing possible 
robot-robot interactions, making the process more like planning single robot 
paths with loose interactions. In future research, it would be interesting to 
exploit the \suo principle further to observe how far we can further minimize 
robot-robot interaction to boost the performance of the system. One immediate 
direction is to add weights to \suoi so that non-optimal single robot paths 
will be generated and gauge the trade-off between optimality loss at the first 
phase and the gain (in computation time and optimality) in the second phase of 
a decouple multi-robot path planner. 
Apart from handling uncertainty in time, we also plan to add mechanisms to treat space uncertainty. 

\textbf{Acknowledgement.} We sincerely thank Joseph W. Durham for many insightful discussions of the work. 


\clearpage
\bibliographystyle{IEEETran}
\bibliography{references}
\clearpage

\appendices
\section{Examples to Show that Vertex, Edge, and Temporal Information are All Essential}\label{app:examples}
\begin{figure}[h!]
    \centering
    \subfloat[Edge info essential]{\includegraphics[width=0.32\linewidth]{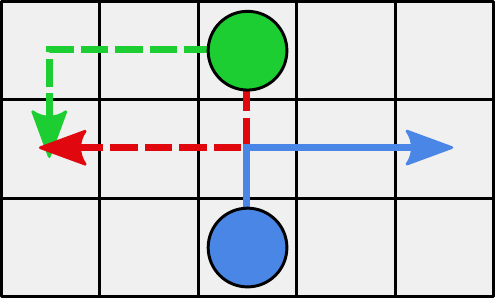}\label{fig:vertex-vs-edge-v}}
    \hspace{1pt}
    \subfloat[Vertex info  essential]{\includegraphics[width=0.32\linewidth]{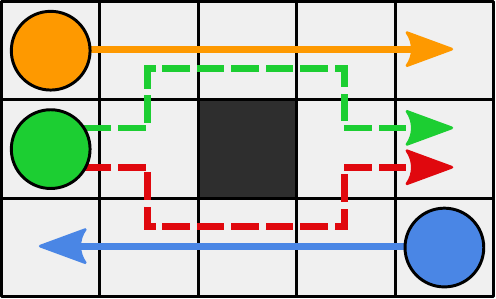}\label{fig:vertex-vs-edge-e}}
    \hspace{1pt}
    \subfloat[Time info  essential]{\includegraphics[width=0.32\linewidth]{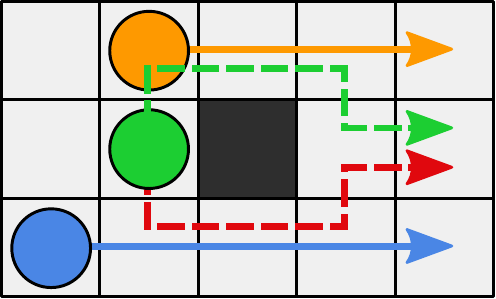}\label{fig:example-temporal}}
    \caption{(a) (b) Two \mpp instances to show vertex and edge information does not dominate each other. 
    (c) An \mpp instance showing that temporal information is helpful
    in selecting a path with fewer potential conflicts. 
    In all scenarios, the solid lines show existing paths that are already planned, 
    The dashed lines are candidate paths for the green robot which we are currently planning. 
    The dark square is an obstacle that cannot be traversed. 
    }
    \label{fig:vertex-vs-edge}
\end{figure}

Fig.~\ref{fig:vertex-vs-edge-v}, \ref{fig:vertex-vs-edge-e} 
illustrate that using only vertex information 
is not strictly better than using only edge information and vice versa. 
In both sub-figures, when planning the path for the green robot, 
the green candidate paths have less conflicts than the red ones.
In Fig.~\ref{fig:vertex-vs-edge-v}, using only edge information cannot tell 
the difference between the green and red paths in terms of path quality 
since the only conflict is between the red and blue paths at the middle {\em vertex}. 
Similarly, in Fig.~\ref{fig:vertex-vs-edge-e}, using only vertex 
information cannot distinguish between the green and red paths since 
they have the same number of vertex conflicts. 
However, the red path also has head-to-head edge conflicts with the blue path. 

Fig.~\ref{fig:example-temporal} highlights the benefit of temporal information. 
Similar to Fig.~\ref{fig:vertex-vs-edge-v}, \ref{fig:vertex-vs-edge-e}, 
vertex and edge information without 
temporal consideration cannot tell the green and red paths apart in 
terms of path quality. But in fact, although the green and orange paths intersect, 
no conflict between them exist because they arrive at same vertices 
at different time steps. 
On the contrary, the blue path will collide with the red path for three 
consecutive time steps at the bottom vertices.

\section{Detailed Proof of \suoi Properties}\label{app:proof}

Proof of Lemma~\ref{lemma:shortest}:
\begin{proof}
We denote the shortest path length for robot $i$ as $|P_i^*|$, 
and a path returned by the described A* algorithm as $P_i$. 
Suppose $|P_i| > |P_i^*|$, then there must exist a vertex $v$ on $P_i^*$ 
which is explored (adjacent to some vertex in $P_i$) but not expanded. 
We further denote $G(v)$ as the cost-to-come for $v$ and 
$F(v) = G(v) + \mathcal H(v)$ as the priority value of $v$ in A* open list. 
We have
\begin{align*}
      & \, F(g_i) - F(v) \\
    = & \, (G(g_i) + H_{\suoi}(g_i)) - (G(v) + H_{\text{short}}(v, g_i) + H_{\suoi}(v)) \\
    = & \, G(g_i) - (G(v) + H_{\text{short}}(u, g_i)) - (H_{\suoi}(v) - H_{\suoi}(g_i)) \\
    > & \, T_i - T_i^* - 1 \geq 0, 
\end{align*}
which indicates that vertex $v$ must be explored earlier than $g_i$. 
We find a contradiction.
\end{proof}

Proof of Lemma~\ref{lemma:minimum}:
\begin{proof}
Lemma~\ref{lemma:shortest} implies that that all vertex expanded during A* search 
are on some shortest paths from $s_i$ to $g_i$. 
Reusing the definition of $F$ from the proof of Lemma~\ref{lemma:shortest}, 
for an expanded vertex $v$, we have 
\[F(v) = G(v) + H_{\text{short}}(v, g_i) + H_{\suoi}(v) = T_i^* + H_{\suoi}(v).\] 
Define $Q_i$ as an arbitrary shortest path for robot $i$. 
Denote $\text{Im}(P_i)$ as the set of all vertices traversed by $P_i$. 
By the searching property of the A* algorithm, we have
\begin{align*}
    \textstyle \max_{v \in \text{Im}(P_i)} F(v) & 
    \leq \textstyle \max_{v \in \text{Im}(Q_i)} F(v), \\
    \textstyle \max_{v \in \text{Im}(P_i)} H_{\suoi}(v) & 
    \leq \textstyle \max_{v \in \text{Im}(Q_i)} H_{\suoi}(v), \\
    C_{\text{single}}(P_i) & \leq C_{\text{single}}(Q_i). \qedhere
\end{align*}
\end{proof}

Proof of Lemma~\ref{lemma:cost-to-go-converge}:
\begin{proof}
It is straightforward that the most conflicted vertex on a single path 
cannot exceed the most conflicted vertex globally, i.e., 
$C_{\text{single}}(P_i) \leq \mathcal C_{\text{single}}$. 
For single robot path planning with result $P_i$, we divide all vertices in $\mathcal G$ 
into two sets: the vertices on $P_i$ and the others. 
$\mathcal C_{\text{single}}$ cannot increase due to vertices in the first set 
since by Lemma~\ref{lemma:minimum}, $P_i$ cannot increase $C_{\text{single}}(P_i)$. 
$\mathcal C_{\text{single}}$ cannot increase due to vertices in the second set 
since the paths for other robots are unchanged. 
\end{proof}

Proof of Lemma~\ref{lemma:cost-to-come-shortest}:
\begin{proof}
$P_i$ is a shortest path because by Lemma~\ref{lemma:occupancy-bound} and the 
definition of $C$, the total accumulated extra cost added by \suoi cannot exceed $1$. 
For minimizing $C_{\text{path}}(P_i)$, note that 
\[\textstyle C_\text{path}(P_i) 
= \frac{n}{\beta_v} \sum_{v \in \text{Im}(P_i)} H_{\suoi}(v), 
\]
so $C_{\text{path}}(P_i)$ is directly associated with the extra \suoi cost, 
which is by definition minimized by A* search. 
\end{proof}

Proof of Lemma~\ref{lemma:cost-to-come-converge}:
\begin{proof}
Due to the fact that collision between two paths is correlated, 
if $C_{\text{path}}(P_i)$ changes, $\sum_{1 \leq j \leq n, j \neq i} C_{\text{path}}(P_j)$ 
will also change by the same amount. 
The proof is completed because each single robot path finding process cannot increase the 
robot's path conflict $C_{\text{path}}(P_i)$. 
\end{proof}

\section{Standalone Evaluation of \suoi}\label{app:evaluation}
We evaluate \suoi using metrics including $\mathcal C_\text{single}$, $\mathcal 
C_\text{path}$ to show that \suoi balances graph utilization and reduces 
path conflicts. The parameters of \suoi are adjusted to the specific metric 
we are optimizing, e.g., when minimizing the number of edge conflicts with 
temporal information, \suoi is set to $\beta_v = 0, \beta_e = 1$ with temporal 
information considered. 
All experiments are performed by planning individual paths for 
$100$ robots with randomly generated starts and goals in $20 \times 10$ grids 
with $5\%$ randomly generated obstacles. 
For plotting, the horizontal axis is the number of \suoi planning iterations $r$; 
$r = 0$ means paths are randomly generated without \suoi. 
The vertical axis is the specific metrics we want to minimize (i.e., lower is 
better); the metrics are normalized to $[0, 1]$. 

\begin{figure}[h!]
    \centering
    \subfloat[\suo as estimated cost-to-go]{\includegraphics{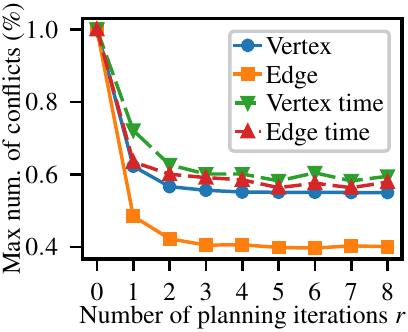}\label{fig:suo-max}}
    \hspace*{1pt}
    \subfloat[Different \suo process orders]{\includegraphics{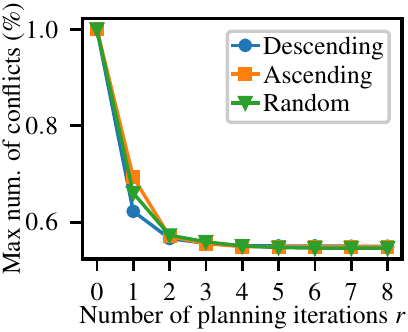}\label{fig:suo-order}}
    \caption{The ratio of the max number of conflicts on a vertex/edge versus 
    the number of \suoi planning iterations. 
    }
    \label{fig:suo-evaluation}
\end{figure}

In Fig.~\ref{fig:suo-max}, we show the metrics with regard to the most utilized graph resource 
when using \suoi as part of estimated cost-to-go. 
For {\em Vertex} entries, the values correspond to the maximum number of times 
a vertex in the graph is traversed. 
For {\em Edge} entries, the values correspond to the maximum number of head-to-head 
conflicts on an edge. 
With the {\em Time} entry, we consider the above metrics on the time domain. 
The plot shows that \suoi significantly reduces the usage of the most conflicted 
graph area. The effect of \suoi improves with increased number of planning 
iterations $r$ and stabilizes after $r \geq 4$, i.e., the reduction diminishes 
after a few iterations.
Using \suoi as part of cost-to-come produces similar result, which is  
omitted here due to limited space. 

In Alg.~\ref{alg:search}, the robots are sorted in the descending distance-to-goal 
order before path planning. In Fig.~\ref{fig:suo-order}, we show the maximum vertex 
collision metric when using descending, ascending, and random order. The plot shows 
that the ordering does not make much difference when $r$ gets large, 
but the descending order is beneficial when $r = 1$.

\end{document}